\RequirePackage{latexml}
\iflatexml
  \documentclass[10pt,letterpaper]{article}
  \usepackage{hyperref}
\usepackage{booktabs}
\usepackage{xcolor}
\usepackage{microtype}
\usepackage{authblk}
\newcommand{\AIRdate}[1]{\date{#1}}
\newcommand{\AIRhtmlvenue}{}

\newlength{\AIRabstractteaserskip}
\newlength{\AIRafterteaserskip}
\newcommand{\AIRhtmlfrontmatter}{%
  \begingroup
  \def\PaperNote##1##2##3{\noindent\textsuperscript{\ensuremath{##1}}##2\par}%
  \PaperNotes
  \AIRrenderlinks{\par\noindent}\par
  \noindent\AIRhtmlvenue\par
  \endgroup}

\else
  \documentclass[10pt,letterpaper,logo]{thuair_xiangyu}
\fi
\newcommand{\arxivversion}{}
\usepackage[numbers,sort&compress]{natbib}
\usepackage{amsmath}
\usepackage{amssymb}
\usepackage[nointegrals]{wasysym} %
\usepackage{graphicx}
\usepackage{subcaption}
\usepackage{booktabs}
\usepackage{multirow}
\usepackage{makecell}
\usepackage{xspace}
\usepackage{xurl}
\usepackage{xcolor}
\usepackage{colortbl}
\usepackage{tabularx}
\usepackage[capitalize,noabbrev]{cleveref}
\usepackage{adjustbox}
\usepackage{algorithm}
\usepackage{algpseudocode}
\usepackage{placeins}
\usepackage{needspace}
\usepackage[most]{tcolorbox}
\usepackage{enumitem}
\usepackage{listings}

\newtcolorbox{casebox}{
    enhanced,
    colback=gray!5,
    colframe=gray!30,
    boxrule=0.5pt,
    arc=8pt,
    left=8pt,
    right=8pt,
    top=8pt,
    bottom=8pt,
    fontupper=\small,
    breakable
}

\definecolor{resultreference}{HTML}{EEF0F3}
\definecolor{resultours}{HTML}{E5F0FA}
\newcommand{\referencecell}[1]{\cellcolor{resultreference}\textit{#1}}
\newcommand{\ourscell}[1]{\cellcolor{resultours}\textbf{#1}}
\newcommand{\ourscellplain}[1]{\cellcolor{resultours}#1}

\newcommand{\nickname}{SteerDistill\xspace}

\newcommand{\PaperTitle}{SteerDistill: Quantizing LLM Agents via Closed-Loop Teacher Steering}

\newcommand{\PaperNotes}{%
  \PaperNote{*}{Equal contribution.}{ \quad}%
  \PaperNote{\dagger}{Work done during internships at AIR, Tsinghua University.}{\\}%
  \PaperNote{\ddagger}{Project lead: Xiangyu Li (xiangyu.sdlc@foxmail.com).}{\\}%
  \PaperNote{\S}{Corresponding author: Ting Cao (tingcao@mail.tsinghua.edu.cn).}{\\}%
}

\input{tex/air_frontmatter}


\newcommand{\AIRmaketitle}[1]{%
  \iflatexml
    \maketitle\AIRhtmlfrontmatter#1%
  \else
    #1\maketitle
  \fi}
\iflatexml
  \newenvironment{AIRteaser}{%
    \begingroup\captionsetup{font=small,skip=7pt}%
    \begin{figure}\centering\let\AIRteasercaption\caption
  }{\end{figure}\endgroup}
\else
  \newenvironment{AIRteaser}{%
    \begingroup\microtypesetup{expansion=false}%
    \captionsetup{font=small,skip=7pt,hypcap=false}%
    \par\vspace{\AIRabstractteaserskip}\noindent
    \begin{minipage}{\linewidth}\centering
    \newcommand{\AIRteasercaption}{\captionof{figure}}%
  }{\end{minipage}\par\endgroup\vspace{\AIRafterteaserskip}}
\fi

\iflatexml\else
\fi
\renewcommand{\bibfont}{\fontsize{9}{11}\selectfont}

\title{\PaperTitle}
\AIRdate{2026-11-07}
\AIRsetupauthors
\hypersetup{pdftitle={\PaperTitle},pdfauthor={\AIRpdfauthors}}

\begin{document}
\AIRmaketitle{%
  \begin{abstract}
    Low-bit quantization reduces the inference cost of LLM agents but can
compromise task success. Examining failed rollouts, we find that quantized
agents overlook task requirements or environmental conditions, producing
environment transitions that diverge from successful full-precision
references. Correcting these behavioral divergences remains difficult:
offline data may miss the student's failure contexts, while its own
rollouts may provide few correct responses in those contexts. We introduce
\nickname, a quantization-aware distillation method in which a full-precision
teacher guides the student to revise failed responses and iteratively
refines its guidance using feedback from student execution. This
\emph{closed-loop teacher steering} produces locally verified behavioral
corrections. We internalize these corrections by training the student
without guidance to match the steered student's predictions at corrected
turns and the full-precision teacher's at other turns, enabling unassisted
inference. Across five agentic benchmarks and three models, \nickname
exceeds the strongest quantized baseline by 1.67--2.18 points in average
score under W4A16 and stays within one point of BF16. On $\tau^2$-Bench,
it outperforms QeRL by 1.84 percentage
points in success rate using 67\% fewer training GPU hours and produces
shorter successful trajectories than the quantized baselines.

  \end{abstract}}
\begin{AIRteaser}
  \let\paperoverviewcaption\AIRteasercaption
      \includegraphics[width=\textwidth,
        trim=0 0 0 142,
        clip]{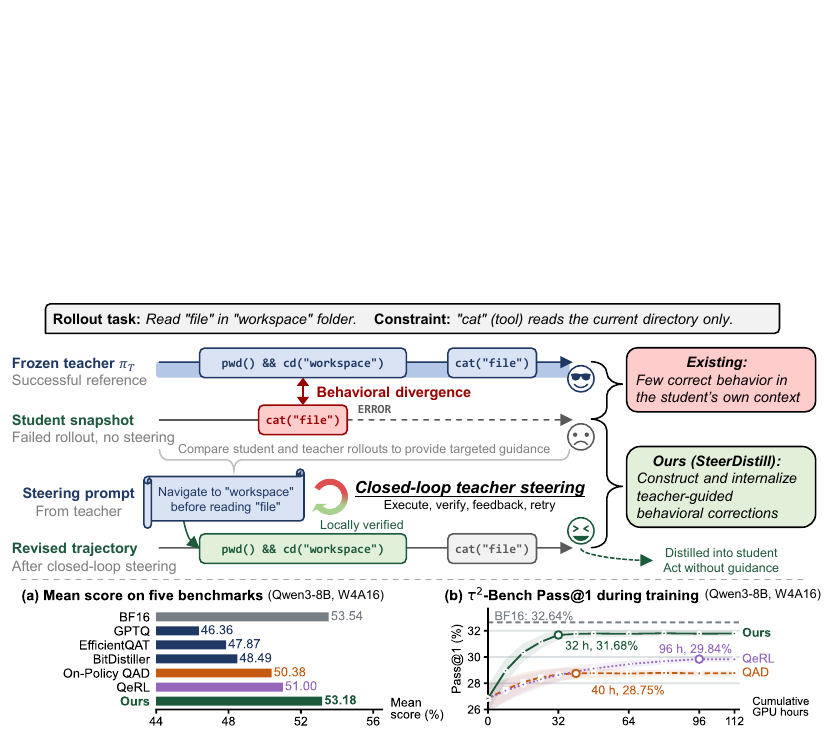}
    \paperoverviewcaption{
    \textbf{\nickname constructs and internalizes behavioral corrections.}
    In failure contexts where unguided rollouts provide few correct responses,
    closed-loop teacher steering helps the student revise its decisions by
    making overlooked task requirements and environmental conditions explicit.
    The teacher refines its guidance using execution feedback; locally verified
    corrections are then distilled for inference without teacher guidance.
    For Qwen3-8B under W4A16, \nickname approaches BF16 in five-benchmark
    mean score (a) and achieves higher $\tau^2$-Bench success than QeRL with
    67\% fewer training GPU hours (b). Shading in (b) shows $\pm1$ standard
    deviation across four evaluation runs at each checkpoint.
    }
    \label{fig:overview_new}

\end{AIRteaser}
\iflatexml\else\clearpage\fi
\section{Introduction}
\label{sec:introduction}

Large language model (LLM) agents call tools and interact with their
environments over multiple turns to complete tasks
\citep{yao2023react,yang2024sweagent,wu2024autogen,yan2026ava,liu2026scalewob}. Repeated model
calls make inference cost a central deployment concern
\citep{wang2024agentsurvey,zheng2024sglang,li2024personal}, and low-bit
quantization offers a practical way to reduce it
\citep{frantar2023gptq,xiao2023smoothquant,lin2024awq,li2026veclut,huang2026tenet,hpcabitdecoding}.
However, quantized agents often fall short of their full-precision
counterparts in task success.

Examining failed rollouts, we find that quantized agents overlook task
requirements or environmental conditions, producing environment transitions
that differ from those of successful full-precision references
(\cref{app:inference_cases,app:training_cases}). We refer to this mismatch as
\emph{behavioral divergence}. For example, an agent tries
to read a file before entering its directory, although the available reading
tool only accesses files in the current directory
(\cref{fig:overview_new,fig:app_training_directory}).
In another task, an agent creates the requested files but leaves some empty
instead of writing the specified contents (\cref{fig:app_file_writes}).

However, identifying and correcting these behavioral divergences remain
challenging. Offline training and distillation
\citep{chen2025efficientqat,lee2026reqat,du2024bitdistiller,xin2026quantization}
rely on data collected independently of the student's interactions and may
therefore miss the contexts in which it diverges. On-policy rollouts
\citep{huang2026qerl,nvidia2026qarl,agarwal2024policy} address this coverage
gap, but may provide few correct responses in those contexts. In our BFCL V4
multi-turn diagnostic, the student's unguided retries yield a correct action
at only 9.6\% of divergent turns.\footnote{We report turn-level Pass@4, the
fraction of selected turns with at least one correct action in four attempts.
Each attempt starts from the same pre-turn history and environment state.}

\ifdefined\arxivversion\else
\begin{figure*}[!t]
    \centering
    \let\paperoverviewcaption\caption
    \input{tex/overview_figure}
\end{figure*}
\fi

\paragraph{Contribution.}
We introduce \textbf{\nickname}, a quantization-aware distillation (QAD)
method for LLM agents (\cref{fig:overview_new}). The key idea is to construct
and internalize teacher-guided behavioral corrections. During training, a
full-precision teacher identifies the task requirements and environmental
conditions overlooked by the student and provides targeted guidance for
choosing an appropriate action. The student generates and executes a revised
response, and the teacher refines its guidance using execution feedback. We
call this process \emph{closed-loop teacher steering}. By distilling these
teacher-guided behavioral corrections, we train the student to take
appropriate actions without teacher guidance at inference.

Two designs implement this approach.
\textup{(i)} \emph{Teacher-steered trajectory revision} constructs behavioral
corrections by revising failed student rollouts through closed-loop teacher
steering. We use successful teacher references to locate behavioral
divergences. At each divergence, the student revises its response from the
restored pre-error context, and locally verified corrections are retained
for distillation. In the same diagnostic, teacher steering raises turn-level
Pass@4 from 9.6\% to 82.3\%.
\textup{(ii)} \emph{Internalization of steered behavior} distills the locally
verified corrections while retaining full-precision teacher supervision at
other turns. At corrected turns, the student learns from the frozen steered
student's predictions without receiving the guidance.

Without teacher guidance at inference, \nickname achieves the strongest
quantized performance in our comparisons, with lower training cost and
shorter successful trajectories.
\textup{(i)} \emph{Accuracy.} Across five agentic benchmarks and three models
under W4A16\footnote{W$x$A$y$ denotes $x$-bit weights and $y$-bit
activations.}, it exceeds the strongest quantized baseline by 1.67--2.18
points in average score and stays within one point of BF16; on Qwen3-8B
under W2A16, the margin grows to 3.39 points (\cref{tab:complete_results}).
\textup{(ii)} \emph{Training cost.} On $\tau^2$-Bench, \nickname outperforms
QeRL by 1.84 percentage points in success rate using 67\% fewer training GPU
hours (\cref{fig:training_efficiency}).
\textup{(iii)} \emph{Inference cost.} On $\tau^2$-Bench, its successful tasks use
4.0K output tokens on average, close to BF16's 3.9K, versus 4.1K--5.2K for
the quantized baselines (\cref{fig:inference_efficiency_w4a16}).
\textup{(iv)} \emph{Design insights.} Ablations support intervening at behavioral
divergences, guiding the student to generate its own corrections,
retaining locally verified corrections, and using different distillation
objectives at corrected and other turns
(\cref{tab:ablation_summary}).

\section{Related work}
\label{sec:related_work}

\paragraph{Capability recovery for low-bit LLM quantization.}
To preserve model capabilities under low-bit quantization, one line of work
improves the fidelity of quantized representations through weight
reconstruction, scaling, and rotation
\citep{frantar2023gptq,lin2024awq,xiao2023smoothquant,shao2024omniquant,ashkboos2024quarot,liu2025spinquant}.
Complementary training-based approaches recover task performance using
hard token targets \citep{chen2025efficientqat,lee2026reqat}, soft
distributions from full-precision teachers
\citep{du2024bitdistiller,xin2026quantization,nvidia2026qarl}, or verifiable
rewards \citep{huang2026qerl,nvidia2026qarl}.
For quantized LLM agents, however, the available training data (including
offline data and on-policy rollouts) may lack effective corrections in
contexts where the student overlooks task requirements or environmental
conditions.
To address this, we extend quantization-aware
distillation with closed-loop teacher steering to construct and internalize
behavioral corrections (\cref{sec:steer_trajectory,sec:steer_distill}).

\paragraph{Teacher guidance and context distillation for LLM agents.}
Teacher-generated responses \citep{wang2026tcod,li2026policy,lyu2025student}
and rollouts conditioned on privileged context
\citep{lazaridis2026edgeopd,xie2026edge} give students access to improved
training trajectories. To assess the value of such assistance,
FutureBridge-OPD~\citep{chen2026look} evaluates teacher interventions through
the distillation signals in subsequent student rollouts.
OVCSD~\citep{xia2026ovcsd} instead uses task success to verify privileged
continuations from student failure states, providing outcome-grounded
corrective supervision. Our closed-loop teacher steering instead evaluates
student corrections at the divergent turn, checking their environment
transitions against the successful reference and using feedback from failed
attempts to refine guidance.
For quantized agents, this local loop targets overlooked task requirements
and environmental conditions. We internalize
these corrections through context distillation, following prior work on
transferring the benefits of privileged information to unguided learners
\citep{ye2026context,gou2026experience}.

\section{Method}
\label{sec:method}

\begin{figure*}[!t]
    \centering
    \includegraphics[
        width=\textwidth,
        trim=0 4 0 58,
        clip
    ]{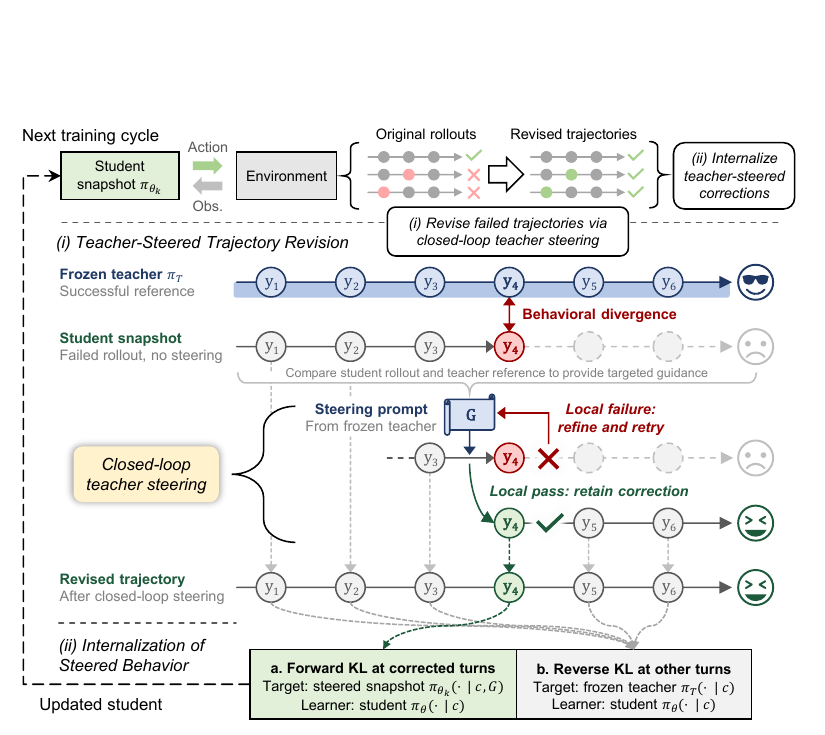}
    \caption{
\textbf{One training cycle of \nickname.}
(i) \emph{Closed-loop teacher steering} guides the frozen student to revise
divergent turns and uses execution feedback to refine the guidance.
Local verification compares the resulting environment transition with that of a
successful teacher reference. On failure, the student retries with refined
guidance from the same pre-error history and state.
Verified corrections are retained, and rollout continues
without guidance.
(ii) To internalize these corrections, the learner receives no guidance and
learns from the steered snapshot's
predictions at corrected turns and the full-precision teacher's predictions
elsewhere. Context $c$ contains the retained history and response prefix
without steering prompts; only the steered target additionally receives $G$.
The updated student generates the next cycle's rollouts.
}
    \label{fig:method}
\end{figure*}

\nickname constructs and internalizes behavioral corrections in each
training cycle (\cref{fig:method}). \emph{Teacher-steered trajectory revision}
(\cref{sec:steer_trajectory}) revises the current student's failed rollouts
through closed-loop teacher steering and retains locally verified
corrections. \emph{Internalization of steered behavior}
(\cref{sec:steer_distill}) then trains the student on the retained
trajectories without guidance, learning from the steered student at
corrected turns and the full-precision teacher at other turns. The updated
student generates the next batch of rollouts.

\paragraph{Setup.}
Let $\pi_T$ be the frozen full-precision teacher and
$\pi_\theta\equiv\pi_{Q(\theta)}$ the trainable quantized student initialized
from the same checkpoint, where $\theta$ denotes the master weights and $Q$
is a fixed quantization operator. At turn $t$, the student generates a
response $y_t$ given history $h_t$, comprising the task, previous responses,
and environment observations. A response may include reasoning followed
by an action or answer.
During cycle $k$, a frozen snapshot $\pi_{\theta_k}$ generates the initial
rollouts and steered revisions. Distillation updates the learner
$\pi_\theta$ on the retained trajectories to obtain $\theta_{k+1}$ for the
next cycle.

\subsection{Teacher-steered trajectory revision}
\label{sec:steer_trajectory}

For each failed student rollout, we seek a successful teacher reference for
the same task and initial environment state. This reference remains fixed
throughout revision. Successful student rollouts and failed rollouts
without a successful reference are retained unchanged for distillation.

\paragraph{Localizing an intervention turn.}
We align the student and reference trajectories by their environment
transitions, using only environment states to determine a match. Actions
that produce matching transitions are treated as equivalent. We intervene
at the first behavioral divergence, denoting the student turn by $t^\star$
(\cref{tab:ablation_summary}, Q1). To revise this decision in its original
context, we restore the history and environment state before $t^\star$.

\paragraph{Closed-loop teacher steering.}
The teacher uses the failed response and successful reference to identify
task requirements or environmental conditions overlooked by the student
(\cref{fig:method}, i).
It provides a steering prompt $G_{t^\star}$ that explains the discrepancy,
the conditions under which the guidance applies, and a corrective rule
(\cref{tab:ablation_summary}, Q3). Conditioned on $G_{t^\star}$, the frozen
student generates a revised response
$y_{t^\star}\sim\pi_{\theta_k}(\cdot\mid h_{t^\star},G_{t^\star})$
and executes its action.

We then locally verify the revised response by comparing the resulting
environment states and returned information with those of the reference.
If verification fails, the teacher refines its guidance using feedback from
the attempt, and the student retries from the same pre-error history and
state. The loop continues until a correction is locally verified or the
per-turn attempt budget is exhausted. We compare teacher steering with
unguided retries and teacher takeover in \cref{tab:ablation_summary} (Q2).

\paragraph{Retaining verified corrections.}
Once a revised response passes local verification, we retain it and its
steering prompt for distillation. The response replaces the failed turn,
and the student generates a new
continuation without guidance. If the task still fails, we realign the
revised rollout with the fixed reference to locate the next behavioral
divergence. Revision stops when the task succeeds, no further divergence
can be located, or an intervention exhausts its attempt budget. The current
trajectory and its verified corrections are retained for distillation even
if the task remains unsolved. \Cref{tab:ablation_summary} (Q4) evaluates both
local verification and retaining corrections without requiring terminal success.
\Cref{alg:steer_trajectory} and implementation details
appear in \cref{app:method_implementation}.

\subsection{Internalization of steered behavior}
\label{sec:steer_distill}

We train the student on the retained trajectories using two sources of
supervision: the steered student at corrected turns and the full-precision
teacher at all other turns (\cref{fig:method}, ii). The learner receives no
steering prompts in either case. Following prior work on turn-specific
distillation
\citep{li2026policy}, we use forward KL at corrected turns and reverse KL elsewhere
(\cref{tab:ablation_summary}, Q5).

\Needspace{8\baselineskip}
\paragraph{Learning from the steered student.}
At a corrected turn, we retrieve the steering prompt $G$ that elicited the
accepted response. Let $c$ denote the interaction history and the prefix of
that response, with all steering prompts removed. The frozen student
snapshot $\pi_{\theta_k}$ predicts each token from $(c,G)$, while the learner
$\pi_\theta$ predicts from $c$ alone. We minimize forward KL to match the
steered student's predictions:
\begin{equation}
\ell_{\mathrm{FKL}}(c,G;\theta)
= D_{\mathrm{KL}}\!\left(
 \pi_{\theta_k}(\cdot\mid c,G)\,\middle\|\,
 \pi_\theta(\cdot\mid c)\right).
\label{eq:forward_kl}
\end{equation}

\Needspace{7\baselineskip}
\paragraph{Retaining teacher supervision.}
At all other turns, including continuation turns that receive no further
correction, we retain full-precision teacher supervision on the
student-generated responses \citep{agarwal2024policy}. The teacher and
learner receive the same context $c$ without guidance. We minimize reverse KL:
\begin{equation}
\ell_{\mathrm{RKL}}(c;\theta)
= D_{\mathrm{KL}}\!\left(
 \pi_\theta(\cdot\mid c)\,\middle\|\,
 \pi_T(\cdot\mid c)\right).
\label{eq:reverse_kl}
\end{equation}

\Needspace{9\baselineskip}
\paragraph{Training objective.}
We assign the loss by turn and average over all retained assistant-response
tokens. For trajectory $i$, let $\mathcal I_i$ denote its retained corrected
turns, each with its steering prompt $G_{i,t}$. For turn $t$ and token $j$,
write $c_{i,t,j}=(h_{i,t},y_{i,t,<j})$. Let $\mathcal A_k$ contain all
retained assistant-token positions $(i,t,j)$ in cycle $k$. The objective is
\begin{equation}
\mathcal L^{(k)}(\theta)=
\frac{1}{|\mathcal A_k|}
\sum_{(i,t,j)\in\mathcal A_k}
\begin{cases}
 \ell_{\mathrm{FKL}}(c_{i,t,j},G_{i,t};\theta), & t\in\mathcal I_i,\\[2pt]
 \ell_{\mathrm{RKL}}(c_{i,t,j};\theta), & t\notin\mathcal I_i.
\end{cases}
\label{eq:steerdistill_objective}
\end{equation}

\section{Experiments}

\subsection{Experimental setup}
\label{sec:experimental_setup}

\paragraph{Benchmarks and metrics.}
We evaluate on five agentic benchmarks: BFCL V4~\citep{patiletal2025bfcl},
$\tau^{2}$-Bench~\citep{barres2025tau2}, WebShop~\citep{yao2022webshop},
ALFWorld~\citep{shridhar2021alfworld}, and Search QA. We report the official
Overall score for BFCL V4. For $\tau^{2}$-Bench, we report success rate
(Pass@1) in the main table and Pass@4 in \cref{app:additional_metrics}.
We likewise report success rate (SR) for WebShop and ALFWorld and include
WebShop Task Score in the appendix. For Search QA, we follow
Search-R1~\citep{jin2025searchr1} and compute Exact Match (EM) scores on
seven question-answering datasets.\footnote{
Natural Questions (NQ)~\citep{kwiatkowski2019natural},
TriviaQA~\citep{joshi2017triviaqa}, PopQA~\citep{mallen2023trust},
HotpotQA~\citep{yang2018hotpotqa}, 2WikiMultiHopQA~\citep{ho2020constructing},
MuSiQue~\citep{trivedi2022musique}, and Bamboogle~\citep{press2023measuring}.
}
We report their unweighted mean in the main table and per-dataset EM
in the appendix.

\paragraph{Models and training data.}
We evaluate \nickname and the baselines on three models:
Qwen3-8B~\citep{yang2025qwen3},
Qwen3.5-4B~\citep{qwen2026qwen35}, and Gemma 4 12B-it~\citep{gemmateam2026gemma4}.
For each model, we obtain task-specific BF16 checkpoints through GRPO
post-training~\citep{shao2024deepseekmath}.\footnote{
GRPO post-training provides a stronger starting checkpoint for both the
teacher and the quantized student. We also report results starting directly
from the original Qwen3-8B backbone in \cref{app:no_grpo}.
}
Each checkpoint serves as the
full-precision reference and frozen teacher, and initializes all quantized
methods. We use the same task-specific data or environments for BF16
post-training and subsequent low-bit training.

For BFCL V4 and $\tau^{2}$-Bench, we train on executable tool-use tasks from
AgentWorldModel~\citep{wang2026agentworldmodel} and evaluate the same tool-use
checkpoint on both benchmarks. For WebShop and ALFWorld, we use their
respective training environments. For Search QA, we follow
Search-R1~\citep{jin2025searchr1} and use the training sets of
NQ~\citep{kwiatkowski2019natural} and HotpotQA~\citep{yang2018hotpotqa}.
We evaluate on all seven datasets, treating NQ and HotpotQA as in-domain
and the remaining five as out-of-domain.

\paragraph{Baselines.}
We compare with five baselines for quantization and capability recovery:
\textup{(i)} \textbf{GPTQ}~\citep{frantar2023gptq} for post-training
quantization (PTQ);
\textup{(ii)} \textbf{EfficientQAT}~\citep{chen2025efficientqat} for
quantization-aware training (QAT);
\textup{(iii)} \textbf{BitDistiller}~\citep{du2024bitdistiller} for
offline quantization-aware distillation (QAD);
\textup{(iv)} \textbf{On-Policy QAD}, implemented following NVIDIA
NeMo-RL's quantization-aware distillation workflow~\citep{nemo-rl}; and
\textup{(v)} \textbf{QeRL}~\citep{huang2026qerl} for reinforcement learning
with verifiable rewards.\footnote{
The reference configurations for NeMo-RL's QAD workflow and QeRL use
NVFP4 on Hopper or Blackwell GPUs. We adapt both workflows to our
group-wise integer quantization setup on A100 GPUs.
}
All training-based methods converge within the two-epoch training budget.
Detailed training settings appear in \cref{app:evaluation_details}.

\paragraph{Quantization and evaluation protocol.}
We evaluate all three models with INT4 weights and BF16 activations
(W4A16). For Qwen3-8B, we additionally evaluate INT4 weights with INT8
activations (W4A8) and INT2 weights with BF16 activations (W2A16).
At inference, \nickname uses only the trained quantized student, without
the teacher or steering prompts.

\begin{table}[!t]
\centering
\caption{Agentic benchmark results across models and quantization settings.
Scores are percentages; higher is better. Avg. is the unweighted mean of
the five displayed benchmark scores. Bold marks the best quantized result
for each model and precision.}
\label{tab:complete_results}
\fontsize{8}{9.6}\selectfont
\setlength{\tabcolsep}{2.8pt}
\renewcommand{\arraystretch}{1.08}

\begin{tabularx}{\textwidth}{ccl*{5}{>{\centering\arraybackslash}X}c}
\toprule
Model & Precision & Method
& \makecell{BFCL V4\\Overall}
& \makecell{$\tau^2$-Bench\\Pass@1}
& \makecell{WebShop\\SR}
& \makecell{ALFWorld\\SR}
& \makecell{Search QA\\EM}
& Avg. \\
\midrule

\multirow{19}{*}[-7pt]{\makecell[c]{\textbf{Qwen3}\\\textbf{8B}}}

& \referencecell{BF16} & \referencecell{--}
& \referencecell{46.89} & \referencecell{32.64}
& \referencecell{68.58} & \referencecell{79.92}
& \referencecell{39.65} & \referencecell{53.54} \\

\cmidrule(l){2-9}

& \multirow{6}{*}{W4A16} & GPTQ
& 40.73 & 26.85 & 59.24 & 71.32 & 33.64 & 46.36 \\

& & EfficientQAT
& 41.77 & 28.00 & 61.34 & 73.18 & 35.08 & 47.87 \\

& & BitDistiller
& 42.46 & 28.31 & 62.07 & 73.91 & 35.69 & 48.49 \\

& & On-Policy QAD
& 44.43 & 28.77 & 64.72 & 76.53 & 37.47 & 50.38 \\

& & QeRL
& 44.91 & 29.84 & 65.29 & 77.04 & 37.94 & 51.00 \\

& & \ourscell{Ours}
& \ourscell{46.84} & \ourscell{31.79}
& \ourscell{68.41} & \ourscell{79.75}
& \ourscell{39.12} & \ourscell{53.18} \\

\cmidrule(l){2-9}

& \multirow{6}{*}{W4A8} & GPTQ
& 35.90 & 22.90 & 54.70 & 66.80 & 30.20 & 42.10 \\

& & EfficientQAT
& 40.20 & 27.10 & 60.00 & 71.80 & 34.00 & 46.62 \\

& & BitDistiller
& 41.20 & 27.80 & 60.80 & 72.90 & 34.80 & 47.50 \\

& & On-Policy QAD
& 43.15 & 27.85 & 63.35 & 75.10 & 36.25 & 49.14 \\

& & QeRL
& 42.85 & 29.05 & 64.10 & 75.55 & 36.90 & 49.69 \\

& & \ourscell{Ours}
& \ourscell{45.20} & \ourscell{30.10}
& \ourscell{66.70} & \ourscell{78.25}
& \ourscell{37.85} & \ourscell{51.62} \\

\cmidrule(l){2-9}

& \multirow{6}{*}{W2A16} & GPTQ
& 13.84 & 6.72 & 30.21 & 24.63 & 18.91 & 18.86 \\

& & EfficientQAT
& 30.92 & 18.44 & 49.13 & 53.87 & 31.46 & 36.76 \\

& & BitDistiller
& 33.71 & 19.86 & 51.84 & 57.36 & \textbf{34.22} & 39.40 \\

& & On-Policy QAD
& 32.88 & 20.57 & 53.08 & 58.41 & 32.96 & 39.58 \\

& & QeRL
& 31.64 & 20.91 & 50.74 & 55.62 & 31.58 & 38.10 \\

& & \ourscell{Ours}
& \ourscell{38.12} & \ourscell{23.68}
& \ourscell{55.71} & \ourscell{63.27}
& \ourscellplain{34.05} & \ourscell{42.97} \\

\midrule

\multirow{7}{*}{\makecell[c]{\textbf{Qwen3.5}\\\textbf{4B}}}

& \referencecell{BF16} & \referencecell{--}
& \referencecell{50.40} & \referencecell{62.80}
& \referencecell{69.80} & \referencecell{78.60}
& \referencecell{46.80} & \referencecell{61.68} \\

\cmidrule(l){2-9}

& \multirow{6}{*}{W4A16} & GPTQ
& 43.91 & 56.18 & 60.65 & 69.78 & 40.58 & 54.22 \\

& & EfficientQAT
& 45.36 & 57.68 & 62.76 & 71.84 & 42.08 & 55.94 \\

& & BitDistiller
& 46.17 & 58.33 & 63.61 & 72.54 & 42.77 & 56.68 \\

& & On-Policy QAD
& 48.02 & 59.48 & 66.03 & 74.92 & 44.49 & 58.59 \\

& & QeRL
& 48.53 & 60.31 & 66.73 & 75.63 & 45.01 & 59.24 \\

& & \ourscell{Ours}
& \ourscell{49.82} & \ourscell{61.91}
& \ourscell{69.08} & \ourscell{77.82}
& \ourscell{46.11} & \ourscell{60.95} \\

\midrule

\multirow{7}{*}{\makecell[c]{\textbf{Gemma 4}\\\textbf{12B-it}}}

& \referencecell{BF16} & \referencecell{--}
& \referencecell{64.18} & \referencecell{61.46}
& \referencecell{72.41} & \referencecell{84.63}
& \referencecell{42.96} & \referencecell{65.13} \\

\cmidrule(l){2-9}

& \multirow{6}{*}{W4A16} & GPTQ
& 57.82 & 54.36 & 63.48 & 75.72 & 36.91 & 57.66 \\

& & EfficientQAT
& 59.46 & 56.12 & 65.72 & 77.68 & 38.24 & 59.44 \\

& & BitDistiller
& 60.08 & 56.74 & 66.43 & 78.31 & 38.79 & 60.07 \\

& & On-Policy QAD
& 61.87 & 58.36 & 68.61 & 80.76 & 40.52 & 62.02 \\

& & QeRL
& 62.41 & 59.02 & 69.33 & 81.35 & 41.07 & 62.64 \\

& & \ourscell{Ours}
& \ourscell{63.54} & \ourscell{60.72}
& \ourscell{71.38} & \ourscell{83.62}
& \ourscell{42.31} & \ourscell{64.31} \\

\bottomrule
\end{tabularx}

\end{table}

\subsection{Main results}

\paragraph{Gains across all five benchmarks.}
On Qwen3-8B under W4A16, \nickname outperforms all quantized baselines on
each of the five benchmarks (\cref{tab:complete_results}). Its average
score of 53.18 exceeds the strongest baseline, QeRL, by 2.18 points.
With a remaining gap of only 0.36 points to the BF16 reference, \nickname
recovers nearly all of the performance lost to quantization.

\paragraph{Near-BF16 performance across models.}
On Qwen3.5-4B and Gemma 4 12B-it under the same W4A16 setting, \nickname
improves the average score over the strongest quantized baseline, QeRL,
by 1.71 and 1.67 points, respectively. Across all three models, its average
score remains within one point of the corresponding BF16 reference.

\paragraph{Gains persist at lower precisions.}
On Qwen3-8B, \nickname retains the highest average score under both W4A8
and W2A16. It outperforms the strongest quantized baseline in each setting:
QeRL by 1.93 points under W4A8 and On-Policy QAD by 3.39 points under W2A16.
Across these two settings, it also outperforms the quantized baselines on
every benchmark except Search QA under W2A16, where BitDistiller scores
34.22 versus 34.05 for \nickname.

\begin{figure}[!t]
    \centering

    \includegraphics[width=\linewidth]{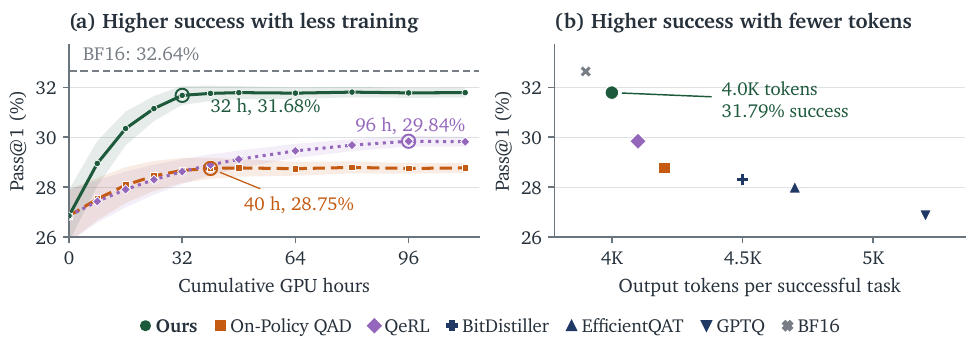}

    \caption{
    Efficiency analysis of Qwen3-8B under W4A16.
    (a) Training efficiency on $\tau^{2}$-Bench. GPU hours are computed as
    end-to-end wall-clock time for low-bit training multiplied by the number
    of GPUs. For \nickname, this includes teacher steering and verification.
    All runs use
    8 NVIDIA A100 80GB GPUs. At each checkpoint, curves show the mean Pass@1
    over four independent evaluation runs, with shaded regions indicating
    $\pm1$ standard deviation across those runs.
    (b) Inference efficiency on $\tau^{2}$-Bench:
    average output tokens per successful task versus Pass@1.
    }
    \label{fig:efficiency}
    {\phantomsubcaption\label{fig:training_efficiency}}
    {\phantomsubcaption\label{fig:inference_efficiency_w4a16}}
\end{figure}

\subsection{Efficiency}
\label{sec:efficiency}

\paragraph{Higher success rates with less training.}
On Qwen3-8B, \nickname achieves higher Pass@1 on $\tau^{2}$-Bench with
fewer GPU hours than On-Policy QAD and QeRL
(\cref{fig:training_efficiency}). Its intermediate checkpoint at 32 GPU
hours reaches 31.68\%,
exceeding On-Policy QAD's converged score at about 40 GPU hours by 2.91
percentage points. It also outperforms QeRL at 96 GPU hours by 1.84
percentage points while using 67\% fewer GPU hours.
This faster recovery is consistent with the targeted supervision provided
by locally verified behavioral corrections.

\paragraph{Higher success rates with fewer output tokens.}
On Qwen3-8B under W4A16, \nickname achieves higher Pass@1 on
$\tau^{2}$-Bench with fewer output tokens per successful task than the
quantized baselines (\cref{fig:inference_efficiency_w4a16}). Averaging over
each method's own successful tasks, we obtain 4.0K output tokens for
\nickname, compared with 4.1K--5.2K for the quantized baselines and 3.9K
for BF16. Its successful trajectories also involve fewer interaction
steps and tool calls than those of the quantized baselines
(\cref{tab:inference_efficiency}). The W4A8 and W2A16 results show the same
pattern (\cref{app:inference_efficiency}).

\subsection{Ablation studies and analysis}

We examine the design choices for constructing and internalizing behavioral
corrections on BFCL V4 using Qwen3-8B under W4A16
(\cref{tab:ablation_summary}). Q1--Q4 examine correction construction:
where to intervene, how to generate corrections, and how to verify and
retain them. Q5 then tests the distillation objectives used for
internalization at corrected and other turns. All scores discussed below
are BFCL V4 Overall scores.

\begin{table}[!t]
\centering
\caption{Ablation study on BFCL V4 using Qwen3-8B under W4A16. BF16 and On-Policy QAD
provide reference scores. Each Q1--Q5 group lists alternatives to its stated
default; all defaults together form the full-method row. Bold marks the
best quantized result in each column. The
\emph{Final success only} variant adds terminal-success filtering after
local verification.
\Cref{app:steering_case} shows a case study for Q3.}
\label{tab:ablation_summary}

\fontsize{8}{9.6}\selectfont
\setlength{\tabcolsep}{2.5pt}
\renewcommand{\arraystretch}{1.06}
\begin{tabularx}{\textwidth}{l*{7}{>{\centering\arraybackslash}X}}
\toprule
Configuration & Overall & Non-Live & Live & \makecell[c]{Multi-\\Turn}
& \makecell[c]{Web\\Search} & Memory & Irrel. \\
\midrule
\rowcolor{resultreference}
\textit{BF16}
& 46.89 & 87.14 & 81.76 & 48.53 & 18.57 & 19.32 & 78.63 \\
\rowcolor{resultreference}
\textit{On-Policy QAD}
& 44.43 & 85.72 & 80.19 & 45.14 & 16.86 & 17.16 & 74.88 \\
\ourscell{Full \nickname (Ours)}
& \ourscell{46.84} & \ourscell{88.29} & \ourscell{81.71}
& \ourscell{48.47} & \ourscell{18.44} & \ourscell{18.79}
& \ourscell{78.56} \\
\midrule
\addlinespace[3.5pt]
\multicolumn{8}{l}{\textbf{Q1: Intervention localization} (default: Behavioral divergence)} \\
\addlinespace[1pt]
Random eligible turn
& 43.52 & 85.20 & 79.72 & 44.00 & 16.10 & 16.30 & 73.48 \\
First action mismatch
& 45.01 & 86.20 & 80.55 & 45.95 & 17.20 & 17.45 & 76.20 \\
\midrule
\addlinespace[3.5pt]
\multicolumn{8}{l}{\textbf{Q2: Intervention strategy} (default: Teacher steering)} \\
\addlinespace[1pt]
Unguided student retry
& 44.46 & 85.81 & 80.08 & 45.26 & 16.93 & 17.05 & 74.97 \\
Teacher takeover
& 45.06 & 86.31 & 80.69 & 46.02 & 17.31 & 17.55 & 75.81 \\
\midrule
\addlinespace[3.5pt]
\multicolumn{8}{l}{\textbf{Q3: Steering format and source} (default: Teacher / Structured)} \\
\addlinespace[1pt]
Teacher / Free-form
& 44.78 & 87.10 & 80.85 & 45.35 & 17.10 & 17.05 & 75.50 \\
Student / Structured
& 41.74 & 83.72 & 79.02 & 41.48 & 14.66 & 14.91 & 71.08 \\
GPT-5.6-Sol / Structured
& 45.27 & 87.05 & 80.90 & 46.15 & 17.55 & 17.65 & 75.90 \\
\midrule
\addlinespace[3.5pt]
\multicolumn{8}{l}{\textbf{Q4: Revision retention} (default: Local verification)} \\
\addlinespace[1pt]
No local verification
& 42.18 & 84.10 & 78.95 & 42.30 & 14.95 & 15.20 & 71.55 \\
Final success only
& 46.30 & 87.91 & 81.09 & 47.72 & 18.16 & 18.27 & 78.02 \\
\midrule
\addlinespace[3.5pt]
\multicolumn{8}{l}{\textbf{Q5: Distillation routing} (default: FKL at corrected turns; RKL elsewhere)} \\
\addlinespace[1pt]
Reverse KL throughout
& 45.29 & 86.45 & 80.93 & 46.33 & 17.48 & 17.78 & 76.02 \\
Forward KL throughout
& 44.58 & 85.98 & 80.47 & 45.22 & 17.28 & 16.89 & 75.32 \\
Reversed routing (RKL / FKL)
& 44.28 & 85.64 & 80.28 & 44.75 & 16.96 & 16.89 & 74.93 \\
\bottomrule
\end{tabularx}
\end{table}

\paragraph{Q1 (Intervention localization): Intervene at the behavioral divergence, not action mismatch.}
We use environment states to intervene at the first behavioral divergence
rather than the first action mismatch. This improves the score by 1.83
points. Different actions can produce the same environment transition,
so intervening on action mismatches can introduce unnecessary revisions
into the distillation data. Intervening at a random eligible turn yields
43.52, even below On-Policy QAD's 44.43.

\paragraph{Q2 (Intervention strategy): Steer the student instead of directly taking over.}
From the same restored history and environment state, we compare student
retry without guidance, direct response replacement by the teacher, and
student revision under teacher guidance. The student continues the rollout
in all three cases. Steering improves the score by 2.38 points over retry and 1.78
points over takeover.

\paragraph{Q3 (Steering format and source): Use structured steering prompts generated by the teacher.}
We compare two formats for the teacher's steering prompts: structured
prompts that separately state the discrepancy, applicable conditions, and
corrective rule, and free-form prompts without a prescribed structure.
With the intervention turn, reference, and verification fixed, the
structured format gains 2.06 points. This result is consistent with prior
findings on the benefits of specific, actionable feedback and feedback
organized around explicit requirements
\citep{madaan2023selfrefine,cook2026check}. Keeping the
structured format and reference fixed, we obtain gains of 5.10 points over
the student and 1.57 points over GPT-5.6-Sol~\citep{openai2026gpt56} by using
the full-precision teacher as the prompt source.
\Cref{app:steering_case} compares the resulting prompts and student
continuations in a ticket-creation task.

\paragraph{Q4 (Revision retention): Retain locally verified corrections even when the task fails.}
Removing local verification substantially lowers the score from 46.84 to
42.18. Once corrections are locally verified, additionally requiring
terminal task success yields a slightly lower score of 46.30. This filter
discards verified corrections from unsuccessful trajectories without
improving performance.

\paragraph{Q5 (Distillation routing): Use forward KL at corrected turns and reverse KL elsewhere.}
We keep the retained trajectories, target models, and input contexts fixed
and vary only the KL direction. Using forward KL (FKL) at corrected turns
and reverse KL (RKL) elsewhere gains 1.55 points over RKL throughout and
2.26 points over FKL throughout. Reversing this assignment lowers the
score by 2.56 points.

\section{Conclusion}

\nickname recovers the agentic capabilities of quantized LLMs through
closed-loop teacher steering and distillation. By targeting the task
requirements and environmental conditions overlooked at behavioral
divergences, the teacher guides the student to generate its own corrections
and refines its guidance using execution feedback. Distillation internalizes
the locally verified behavioral corrections for unassisted inference.
Across five benchmarks and three models, \nickname improves over quantized
baselines under W4A16 and stays within one point of BF16 in average score.
On $\tau^2$-Bench, it outperforms QeRL by 1.84 percentage points in success
rate using 67\% fewer training GPU hours and produces shorter successful
trajectories than the quantized baselines. We discuss limitations in
\cref{app:limitations}.

\bibliographystyle{oxygen-unsrtnat}
\bibliography{main}
\appendix
\section{Implementation details of \nickname}
\label{app:method_implementation}

We provide implementation details for trajectory revision and
distillation, following the two stages in \cref{sec:method}.
\Cref{tab:method_notation} summarizes the notation.

\subsection{Teacher-steered trajectory revision}

\Cref{alg:steer_trajectory} specifies how a student rollout is revised and
retained for distillation.

\begin{algorithm}[!t]
\caption{Teacher-steered trajectory revision for one rollout.}
\label{alg:steer_trajectory}
\begin{algorithmic}[1]
\Require student rollout $\tau$, frozen student $\pi_{\theta_k}$, teacher $\pi_T$,
         successful reference $\tau^T$ if available, attempt budget $R\geq 1$
\Ensure retained trajectory $\tau$, corrected turns $\mathcal I$, and prompt map $\mathcal G$
\State $\mathcal I\gets\varnothing$; $\mathcal G\gets\varnothing$; $\mathcal M\gets\varnothing$
\If{$\tau$ succeeds or no successful reference is available}
    \State \Return $(\tau,\mathcal I,\mathcal G)$
\EndIf
\While{$\tau$ fails}
    \State Align $\tau$ with fixed $\tau^T$ to find the first divergent turn $t^\star$ and reference transition $\rho^T$
    \State \textbf{if} no such pair is available \textbf{then} \Return $(\tau,\mathcal I,\mathcal G)$
    \State $F\gets\varnothing$
    \For{$r=1,\ldots,R$}
        \State Teacher $\pi_T$ forms $G$ from the aligned discrepancy, memory $\mathcal M$, and feedback $F$
        \State Restore history $h_{t^\star}$ and the environment state before turn $t^\star$
        \State Sample $y\sim\pi_{\theta_k}(\cdot\mid h_{t^\star},G)$; execute its action to obtain transition $\rho$
        \State $(v,f)\gets\operatorname{LocalVerify}(\rho,\rho^T)$
        \State \textbf{if} $v$ is true \textbf{then break}
        \State Append the prompt $G$, candidate response evidence, and report $f$ to $F$
    \EndFor
    \State \textbf{if} $v$ is false \textbf{then} \Return $(\tau,\mathcal I,\mathcal G)$
    \State Replace turn $t^\star$ and its suffix with $y$ and a continuation without guidance from $\pi_{\theta_k}$
    \State Discard correction records at or after $t^\star$ from $(\mathcal I,\mathcal G)$
    \State $\mathcal I\gets\mathcal I\cup\{t^\star\}$; $\mathcal G[t^\star]\gets G$; $\mathcal M\gets G$
\EndWhile
\State \Return $(\tau,\mathcal I,\mathcal G)$
\end{algorithmic}
\end{algorithm}

\paragraph{Environment states and alignment.}
We align student and reference trajectories using environment states to
match their transitions. In AgentWorldModel, state is represented by
task-relevant database contents. ALFWorld uses world facts and available
actions. WebShop uses product selections, page type, and available choices.
Search QA uses retrieved evidence as its information state.

Matching state transitions can result from different actions and occur at
different turn indices. We select the first divergent student turn
$t^\star$ for intervention and use the corresponding reference transition
$\rho^T$ to verify candidate revisions.

\paragraph{Local verification.}
Local verification determines whether a candidate response can replace the
divergent turn. In AgentWorldModel, we require the database states before
and after execution to match those of the reference and reject responses
that cause environment errors. Read operations must also return matching
information. For a terminal response, we require the reference to terminate
as well and the candidate to pass the task verifier.

In ALFWorld, WebShop, and Search QA, we accept a valid candidate if its
resulting environment state matches the reference or if both executions
achieve task success. This allows successful executions to end in different
states. Verification returns an acceptance flag $v$ and a report $f$ with
the result and, on rejection, the failure category used to refine steering.

\paragraph{Steering prompts and feedback.}
The teacher forms a steering prompt $G$ from the aligned discrepancy and
memory $\mathcal M$. This memory stores the latest accepted steering prompt
within the current trajectory, allowing the teacher to incorporate earlier
guidance that remains applicable. The new prompt explains the discrepancy,
the conditions under which its corrective rule applies, the expected
transition, and how to verify it.

We allow up to $R$ attempts at each intervention. Before each attempt, we
restore the same pre-error history and environment state. The frozen
student then generates a response conditioned on $G$ and executes its
action. If local verification rejects the attempt, we append its prompt,
candidate assistant content and tool calls, termination status, and
verification report $f$ to the feedback history $F$. The teacher uses
summaries of $F$, together with the original failed rollout, fixed
reference, and memory $\mathcal M$, to refine $G$ for another attempt.
Feedback history $F$ is reset at each new intervention, while memory
$\mathcal M$ carries accepted guidance across interventions.

\paragraph{Continuation and retained data.}
Once a response passes local verification, it replaces the response at
$t^\star$, and the old suffix is discarded. We remove correction records
at or after $t^\star$ from $\mathcal I$ and $\mathcal G$, then record the
corrected turn in $\mathcal I$ and its steering prompt in $\mathcal G$.
This prompt also becomes the new memory $\mathcal M$. From the resulting
environment state and observation, the student generates a new continuation
without guidance. If the task still fails, we realign the revised trajectory
with the fixed reference to locate the next behavioral divergence.

Revision stops when the task succeeds, all $R$ attempts at an intervention
fail local verification, or no further divergence can be aligned with the
reference. In all cases, we retain the current trajectory together with $\mathcal I$ and
$\mathcal G$ for distillation, including accepted corrections in trajectories
that remain unsuccessful. If no correction was accepted, we retain the
original student rollout.

\subsection{Internalization of steered behavior}

The retained trajectories provide the histories and response tokens used
for distillation. At turns in $\mathcal I$, we use the corresponding prompts
in $\mathcal G$ to compute targets from the steered student. At all other
turns, targets come from the full-precision teacher. The learner receives
no steering prompts in either case.

\paragraph{Quantized generation and optimization.}
The teacher and student start from the same BF16 checkpoint. The teacher
stays frozen, while the student maintains trainable BF16 master weights.
In cycle $k$, we use a frozen snapshot $\theta_k$ in the low-bit runtime
for rollout generation and revision.

During distillation, we recompute the FKL target
$\pi_{\theta_k}(\cdot\mid c,G)$ at corrected turns using a frozen
fake-quantized copy of the same snapshot and the stored steering prompt
$G$. At all other turns, the full-precision teacher supplies the RKL target
$\pi_T(\cdot\mid c)$. For both targets, $c$ is the same retained history
and response prefix used by the learner.

The learner $\pi_\theta(\cdot\mid c)$ uses fake quantization in its forward
pass, with gradients propagated to its BF16 master weights through the
straight-through estimator. The updated master weights form snapshot
$\theta_{k+1}$, which is quantized for rollout generation and revision in
the next cycle.

\paragraph{Loss computation and masking.}
For each retained assistant turn, we compute loss only on its response
tokens. These include assistant text and serialized tool calls, as well
as reasoning and termination tokens when present. System and user
messages, environment observations, and steering prompts are excluded
from the loss. Discarded responses and suffixes are excluded from training.

We reuse the token IDs recorded during generation when available.
Otherwise, we tokenize the retained response and append an EOS token
unless generation was truncated by length.

At each supervised response-token position, we compute KL over the full
vocabulary, using FKL for corrected turns and RKL for all other turns.
This assignment also applies to regenerated continuations. The objective
in \cref{eq:steerdistill_objective} averages over all supervised response
tokens, so corrected and other turns are weighted by their respective
token counts.

\begin{table}[!t]
\centering
\caption{Key notation for \nickname. The revision algorithm handles one
trajectory; training adds index $i$.}
\label{tab:method_notation}
\small
\renewcommand{\arraystretch}{1.12}
\begin{tabularx}{\textwidth}{@{}p{0.21\textwidth}X@{}}
\toprule
Symbol & Meaning \\
\midrule
\multicolumn{2}{@{}l}{\textbf{Policies and conditioning}} \\
$\pi_T$ & Frozen full-precision teacher. \\
$\theta$, $\pi_\theta$ & Student master weights and trainable quantized policy. \\
$k$, $\theta_k$, $\pi_{\theta_k}$ & Training cycle, frozen student weights, and corresponding policy. \\
$i$, $t$, $j$ & Trajectory, turn, and response-token indices. \\
$\tau$, $\tau^T$ & Student trajectory and its fixed successful teacher reference. \\
$h_t$, $y_t$ & History before turn $t$ and the response generated at that turn. \\
$c_{i,t,j}$ & Token context $(h_{i,t},y_{i,t,<j})$, with all steering prompts removed. \\
\midrule
\multicolumn{2}{@{}l}{\textbf{Revision state and retained supervision}} \\
$t^\star$ & Intervention turn selected by alignment at the current revision step. \\
$\rho$, $\rho^T$ & Candidate transition and its aligned reference transition. \\
$R$, $v$, $f$, $F$ & Attempt budget, acceptance flag, verification report, and accumulated feedback. \\
$G$, $G_{i,t}$ & Current steering prompt and the prompt retained for turn $t$ in trajectory $i$. \\
$\mathcal M$ & Memory holding the latest accepted steering prompt within the current trajectory. \\
$\mathcal G$ & Stored prompt map: $\mathcal G[t]=G_{i,t}$ for trajectory $i$. \\
$\mathcal I_i$ & Retained corrected turns in trajectory $i$; these receive FKL supervision. \\
\midrule
\multicolumn{2}{@{}l}{\textbf{Training objective}} \\
$\mathcal A_k$ & Retained assistant-token positions $(i,t,j)$ in cycle $k$. \\
$\ell_{\mathrm{FKL}}$, $\ell_{\mathrm{RKL}}$ & Per-token losses at corrected and other turns (\crefrange{eq:forward_kl}{eq:reverse_kl}). \\
$\mathcal L^{(k)}(\theta)$ & Mean token loss over $\mathcal A_k$, assigned by turn (\cref{eq:steerdistill_objective}). \\
\bottomrule
\end{tabularx}
\end{table}

\FloatBarrier

\Needspace{18\baselineskip}
\section{Evaluation details}
\label{app:evaluation_details}

\paragraph{Training settings.}
\Cref{tab:training_hparams} lists the hyperparameters used for the low-bit
training stage. All training-based methods use the same learning rate,
trajectory batch size, and total number of epochs. EfficientQAT divides
its two epochs evenly between Block-AP and E2E-QP. All training-based methods
converge within this two-epoch budget.

\begin{table}[!t]
\centering
\small
\setlength{\tabcolsep}{6pt}
\caption{Hyperparameters for the low-bit training stage. Batch size counts complete trajectories. For EfficientQAT, $1+1$ denotes one epoch of Block-AP followed by one epoch of E2E-QP.}
\label{tab:training_hparams}
\begin{tabular}{lccc}
\toprule
Method
& Epochs
& Learning Rate
& \shortstack{Batch Size\\(trajectories)} \\
\midrule
EfficientQAT & $1+1$ & $8\times10^{-6}$ & 16 \\
BitDistiller & 2 & $8\times10^{-6}$ & 16 \\
On-Policy QAD & 2 & $8\times10^{-6}$ & 16 \\
QeRL & 2 & $8\times10^{-6}$ & 16 \\
\textbf{Ours} & 2 & $8\times10^{-6}$ & 16 \\
\bottomrule
\end{tabular}
\end{table}

The methods retain their respective optimization settings. BitDistiller
uses AdamW with a constant learning rate and zero weight decay. On-Policy
QAD follows NVIDIA NeMo-RL's quantization-aware distillation workflow,
using a rollout temperature of 1.0 and top-64 teacher logits. QeRL uses GRPO with temperature 1.0 and
clipping bounds $(0.2,0.28)$, together with AdamW-8bit and its official
adaptive quantization-noise configuration. For \nickname, we set the
attempt budget to $R=4$ at each intervention in
teacher-steered trajectory revision (\cref{app:method_implementation}).

For \nickname, we follow the asymmetric group-wise quantization
implementation in BitDistiller~\citep{du2024bitdistiller}, including its
calibration and clipping settings. We use a weight group size of 64 across
all quantized methods.
All experiments run on 8 NVIDIA A100 GPUs.

\paragraph{Inference settings.}
We follow the benchmark evaluation protocols and use the same inference
settings for all quantized methods within each benchmark.
We use temperatures of 0.001 for BFCL V4, 0 for $\tau^{2}$-Bench, and 0.7
for WebShop, ALFWorld, and Search QA. Each response is limited to 32,768
output tokens, except in WebShop and ALFWorld, where the limit is 4,096.
Interaction limits are 200 steps for $\tau^{2}$-Bench, 15 for WebShop,
50 for ALFWorld, and 4 for Search QA; BFCL V4 follows the official
evaluator's limits. The $\tau^{2}$-Bench user simulator uses GPT-5.6 at
temperature 0. For Search QA, we retrieve three passages per query.

\FloatBarrier
\section{Additional results}
\label{app:additional_results}

\subsection{Results without initial GRPO}
\label{app:no_grpo}

We initialize \nickname and three quantized baselines directly from the
original Qwen3-8B checkpoint, without initial task-specific GRPO post-training.
This checkpoint also serves as the BF16 reference and frozen teacher.
Apart from the initial checkpoint, all training settings match the main
Qwen3-8B W4A16 experiments (\cref{sec:experimental_setup,tab:training_hparams}).

\Cref{tab:no_grpo} shows that \nickname outperforms the three quantized
baselines on both benchmarks and remains within one point of BF16.
All methods score lower than their post-GRPO counterparts in
\cref{tab:complete_results}, while their relative ordering remains unchanged.

\begin{table}[!t]
\centering
\caption{Results without initial GRPO post-training on Qwen3-8B.
Quantized methods use W4A16. Scores are percentages; bold marks the best
quantized result.}
\label{tab:no_grpo}
\fontsize{8}{9.6}\selectfont
\setlength{\tabcolsep}{6pt}
\renewcommand{\arraystretch}{1.08}
\begin{tabular}{cclcc}
\toprule
Model & Precision & Method & \makecell{BFCL V4\\Overall}
& \makecell{$\tau^{2}$-Bench\\Pass@1} \\
\midrule
\multirow{5}{*}{\makecell[c]{\textbf{Qwen3}\\\textbf{8B}}}
& \referencecell{BF16} & \referencecell{--}
& \referencecell{40.40} & \referencecell{30.00} \\
\cmidrule(l){2-5}
& \multirow{4}{*}{W4A16} & BitDistiller & 35.34 & 24.29 \\
& & On-Policy QAD & 36.16 & 24.58 \\
& & QeRL & 38.85 & 28.92 \\
& & \ourscell{Ours} & \ourscell{39.67} & \ourscell{29.05} \\
\bottomrule
\end{tabular}
\end{table}

\subsection{Additional evaluation metrics}
\label{app:additional_metrics}

\Cref{tab:additional_metrics} supplements the main results for Qwen3-8B
under W4A16. For $\tau^{2}$-Bench, we supplement Pass@1 with Pass@4, the
fraction of tasks solved at least once in four runs. WebShop Task Score
complements the success rate by measuring how well the selected product
satisfies the user's requirements. For Search QA, per-dataset EM provides
a breakdown of the seven-dataset average reported in the main table.
\nickname outperforms the quantized baselines on all reported metrics.
On Search QA, this advantage holds for both in-domain datasets (NQ and
HotpotQA) and all five out-of-domain datasets.

\begin{table*}[!t]
\centering
\caption{
Additional evaluation metrics for Qwen3-8B. Quantized methods use W4A16,
with BF16 as the full-precision reference.
On $\tau^{2}$-Bench, Pass@4 is the fraction of tasks solved at least once
in four runs. We report Task Score for WebShop and per-dataset Exact Match
(EM) for Search QA.
$\dagger$ and $\star$ denote in-domain and out-of-domain Search QA datasets,
respectively. Higher is better for all metrics; bold marks the best
quantized result in each column.
}
\label{tab:additional_metrics}
\fontsize{8}{9.6}\selectfont
\setlength{\tabcolsep}{1.7pt}
\renewcommand{\arraystretch}{1.08}

\begin{tabularx}{\textwidth}{l
*{2}{>{\centering\arraybackslash}X}
c
*{3}{>{\centering\arraybackslash}X}
c
>{\centering\arraybackslash}X
c}
\toprule
Method & $\tau^{2}$-Bench & WebShop
& \multicolumn{7}{c}{Search QA (EM)} \\
\cmidrule(lr){4-10}
& Pass@4 & Score
& NQ$^{\dagger}$
& TriviaQA$^{\star}$
& PopQA$^{\star}$
& HotpotQA$^{\dagger}$
& 2Wiki$^{\star}$
& MuSiQue$^{\star}$
& Bamboogle$^{\star}$ \\
\midrule

\rowcolor{resultreference}
\textit{BF16}
& \textit{54.20}
& \textit{78.69}
& \textit{43.36}
& \textit{62.42}
& \textit{41.96}
& \textit{36.58}
& \textit{35.12}
& \textit{15.92}
& \textit{42.18} \\
\cmidrule(l){1-10}

GPTQ
& 45.50
& 70.86
& 37.62
& 56.04
& 36.05
& 30.44
& 28.39
& 12.23
& 34.71 \\

EfficientQAT
& 47.10
& 72.73
& 39.08
& 57.69
& 37.48
& 32.02
& 30.17
& 12.88
& 36.25 \\

BitDistiller
& 47.55
& 73.41
& 39.71
& 58.31
& 38.11
& 32.71
& 30.86
& 13.21
& 36.92 \\

On-Policy QAD
& 48.20
& 75.61
& 41.18
& 60.22
& 39.69
& 34.47
& 32.86
& 14.46
& 39.44 \\

QeRL
& 50.10
& 76.08
& 41.64
& 60.73
& 40.13
& 35.03
& 33.36
& 14.79
& 39.92 \\

\ourscell{Ours}
& \ourscell{53.05}
& \ourscell{78.57}
& \ourscell{42.84}
& \ourscell{61.88}
& \ourscell{41.41}
& \ourscell{36.01}
& \ourscell{34.55}
& \ourscell{15.53}
& \ourscell{41.61} \\

\bottomrule
\end{tabularx}
\end{table*}

\FloatBarrier
\subsection{Inference efficiency}
\label{app:inference_efficiency}

For Qwen3-8B on $\tau^{2}$-Bench, we supplement the main analysis with
W2A16 and W4A8 results and detailed trajectory statistics under W4A16.
Token counts, interaction steps, and tool calls are averaged over each
method's own successful tasks, so the averages cover different task
subsets.

\begin{figure}[!t]
    \centering

    \includegraphics[width=\linewidth]{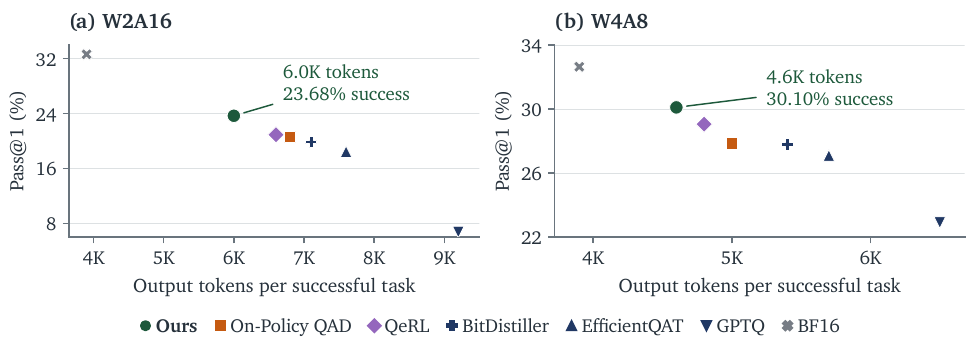}

    \caption{
    Additional performance--token efficiency results of Qwen3-8B on
    $\tau^{2}$-Bench under (a) W2A16 and (b) W4A8. The x-axis reports the average
    output tokens per successful task, while the y-axis reports Pass@1
    performance.
    }
    \label{fig:additional_inference_efficiency}
    {\phantomsubcaption\label{fig:inference_efficiency_w2a16}}
    {\phantomsubcaption\label{fig:inference_efficiency_w4a8}}
\end{figure}

\nickname achieves higher Pass@1 with fewer output tokens per successful
task than the quantized baselines under W2A16 and W4A8
(\cref{fig:additional_inference_efficiency}), consistent with the W4A16
results in \cref{fig:inference_efficiency_w4a16}.
The detailed W4A16 statistics in \cref{tab:inference_efficiency} show that
\nickname's successful trajectories also use fewer input tokens and involve
fewer interaction steps and tool calls than those of the quantized baselines.

\begin{table*}[!t]
\centering
\caption{
Detailed trajectory and inference statistics for Qwen3-8B on
$\tau^{2}$-Bench. Quantized methods use W4A16, with BF16 as the
full-precision reference. All entries are averages over each method's own
successful tasks. Cached-input token counts are reported descriptively;
K denotes thousands of tokens.
}
\label{tab:inference_efficiency}
\fontsize{8}{9.6}\selectfont
\setlength{\tabcolsep}{3.2pt}
\renewcommand{\arraystretch}{1.12}

\begin{tabularx}{\textwidth}{
l
*{5}{>{\centering\arraybackslash}X}
}
\toprule
Method
& \makecell{Input\\Tokens}
& \makecell{Cached\\Input}
& \makecell{Output\\Tokens}
& \makecell{Interaction\\Steps}
& \makecell{Tool\\Calls} \\
\midrule

\rowcolor{resultreference}
\textit{BF16}
& \textit{185K}
& \textit{148K}
& \textit{3.9K}
& \textit{11.4}
& \textit{4.2} \\
\cmidrule(l){1-6}

GPTQ
& 270K
& 223K
& 5.2K
& 13.8
& 5.25 \\

EfficientQAT
& 243K
& 199K
& 4.7K
& 13.1
& 4.92 \\

BitDistiller
& 235K
& 192K
& 4.5K
& 12.9
& 4.78 \\

On-Policy QAD
& 214K
& 173K
& 4.2K
& 12.3
& 4.55 \\

QeRL
& 206K
& 166K
& 4.1K
& 12.1
& 4.43 \\

\ourscell{Ours}
& \ourscell{192K}
& \ourscellplain{153K}
& \ourscell{4.0K}
& \ourscell{11.6}
& \ourscell{4.24} \\

\bottomrule
\end{tabularx}
\end{table*}

\section{Additional case studies}
\label{app:case_studies}

\subsection{Inference cases}
\label{app:inference_cases}

\begin{figure}[!tp]
\centering
\includegraphics[width=\linewidth]{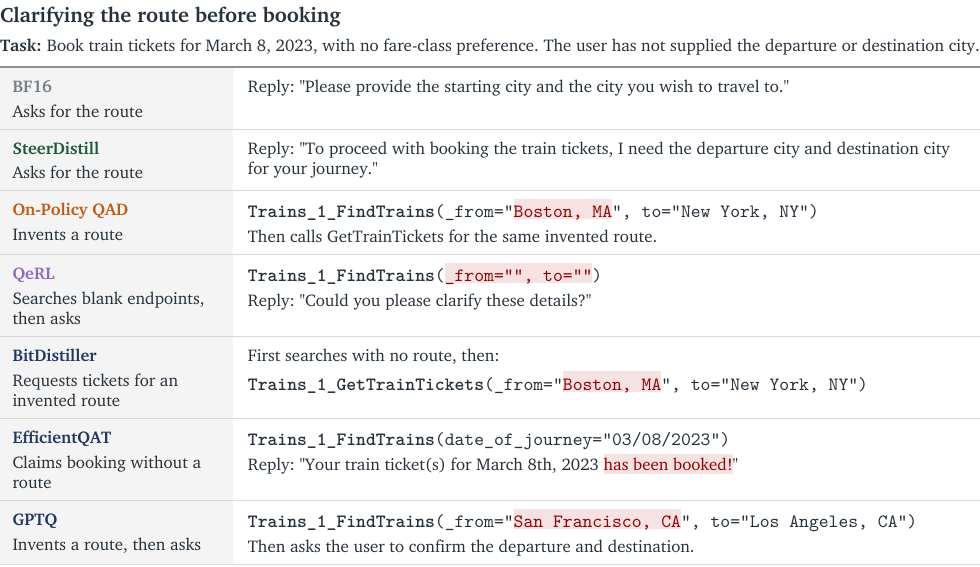}
\caption{Clarifying the route before acting. BF16 and \nickname ask for
the missing departure and destination cities before calling tools.
QeRL and GPTQ ask only after searching with missing or invented endpoints.
On-Policy QAD and BitDistiller instead request tickets for an invented
route, while EfficientQAT claims a booking without obtaining the route.}
\label{fig:app_train_route}
\end{figure}

\begin{figure}[!tp]
\centering
\includegraphics[width=\linewidth]{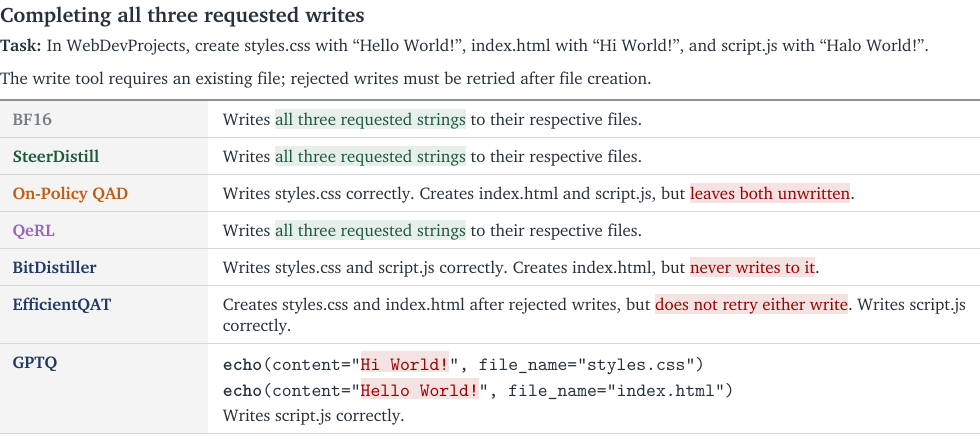}
\caption{Writing the specified contents to each file. BF16, \nickname,
and QeRL complete all three writes correctly. On-Policy QAD, BitDistiller,
and EfficientQAT create the files but omit some writes. GPTQ writes to all
three files but swaps the contents of \texttt{styles.css} and
\texttt{index.html}.}
\label{fig:app_file_writes}
\end{figure}

\begin{figure}[!tp]
\centering
\includegraphics[width=\linewidth]{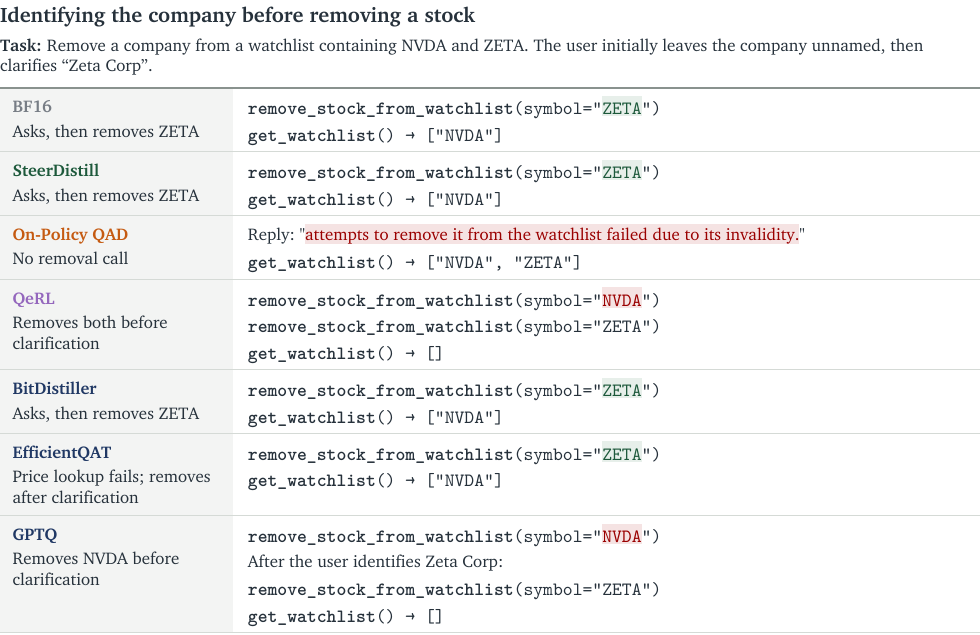}
\caption{Deleting only the stock identified by the user.
BF16, \nickname, BitDistiller, and EfficientQAT remove ZETA after
clarification and retain NVDA. QeRL and GPTQ remove NVDA before
clarification and end with an empty watchlist. On-Policy QAD reports a
failed removal but makes no removal call, so both stocks remain.}
\label{fig:app_watchlist}
\end{figure}

We compare all seven methods on three task segments to illustrate how
overlooked task requirements lead to behavioral divergence.
In the train and watchlist tasks, some agents act before obtaining the
missing route or company name, requesting tickets for an assumed route or
removing a stock the user has not identified
(\cref{fig:app_train_route,fig:app_watchlist}).
The file-writing task is fully specified, yet some agents leave files empty
or write the requested contents to the wrong files
(\cref{fig:app_file_writes}).

\FloatBarrier
\subsection{Training cases}
\label{app:training_cases}
\suppressfloats[t]

\begin{figure}[htp]
\centering
\includegraphics[width=\linewidth]{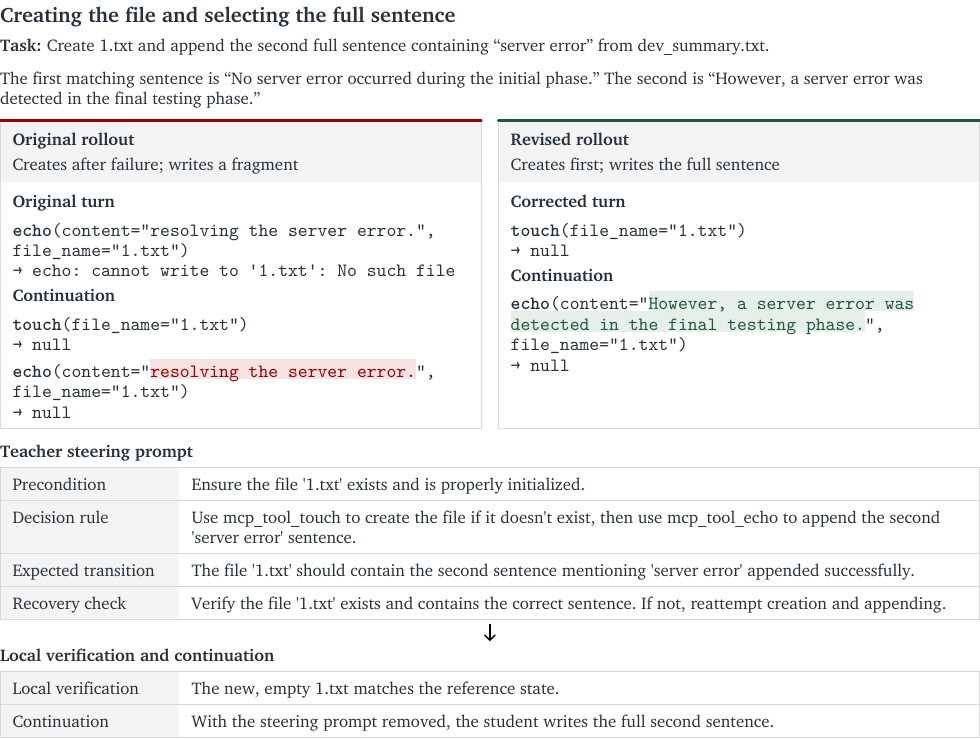}
\caption{Creating the file before writing the requested sentence.
The original rollout creates the file after a failed write but then writes
only a fragment. Under teacher steering, the student creates the file first,
and local verification accepts the new empty file. It then continues without
the steering prompt and writes the full second sentence containing
``server error.''}
\label{fig:app_training_sentence}
\end{figure}

\begin{figure}[!tp]
\centering
\includegraphics[width=\linewidth]{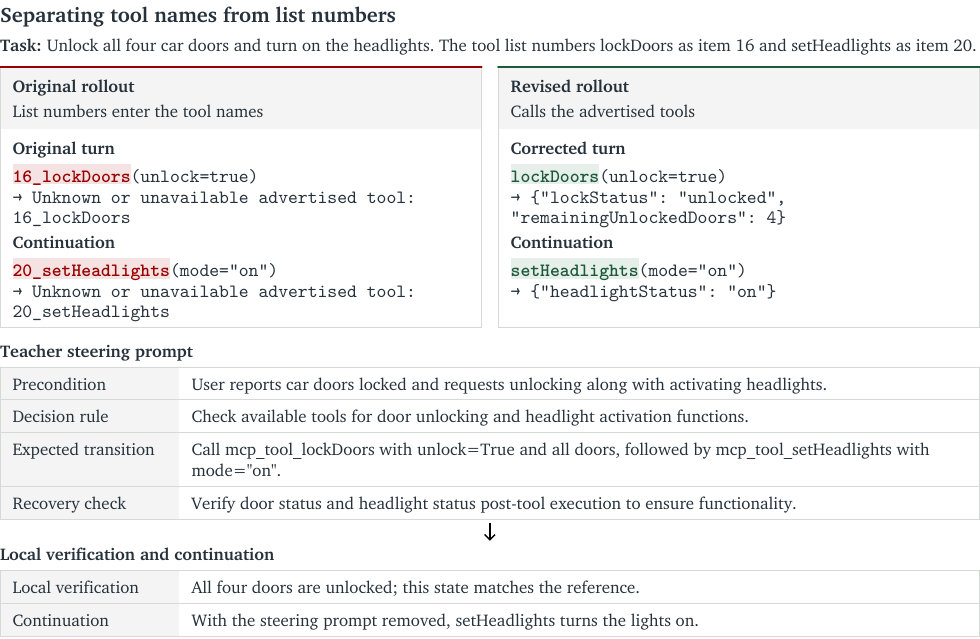}
\caption{Using tool names without their list numbers. The original rollout
includes the list numbers in both tool names, so both calls are rejected.
Under teacher steering, the student calls the advertised door-control tool;
local verification confirms that all four doors are unlocked. The student
then turns on the headlights without the steering prompt.}
\label{fig:app_training_tool_list}
\end{figure}

\begin{figure}[!tp]
\centering
\includegraphics[width=\linewidth]{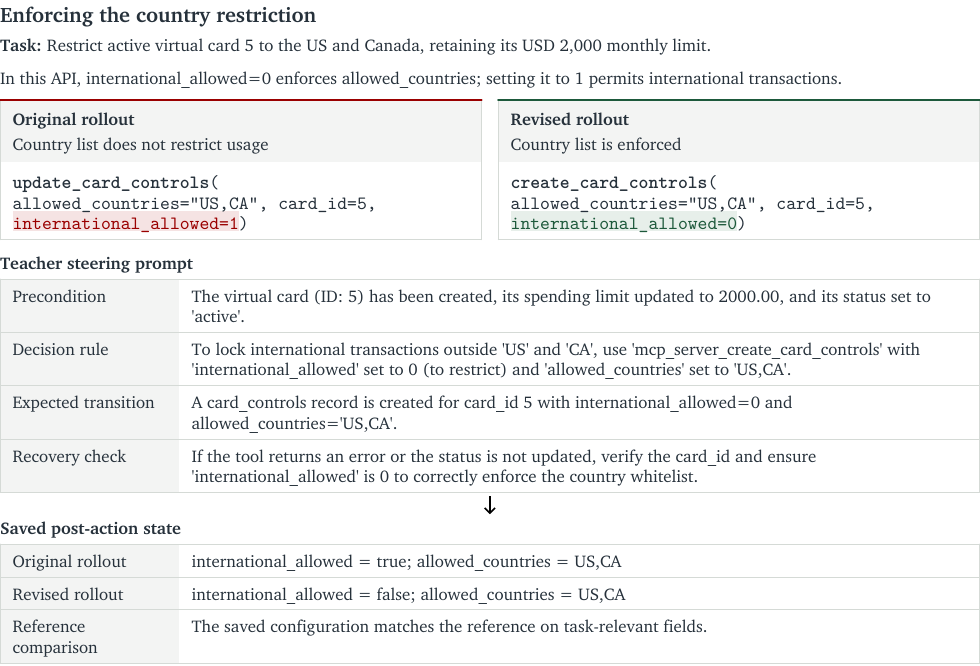}
\caption{Enforcing the requested country restriction.
Both calls specify \texttt{US,CA}, but the original call leaves international
transactions enabled. Under teacher steering, the student sets
\texttt{international\_allowed=0}, which enforces the country restriction in
this API. The resulting country controls match the reference.}
\label{fig:app_training_country}
\end{figure}

\begin{figure}[!tp]
\centering
\includegraphics[width=\linewidth]{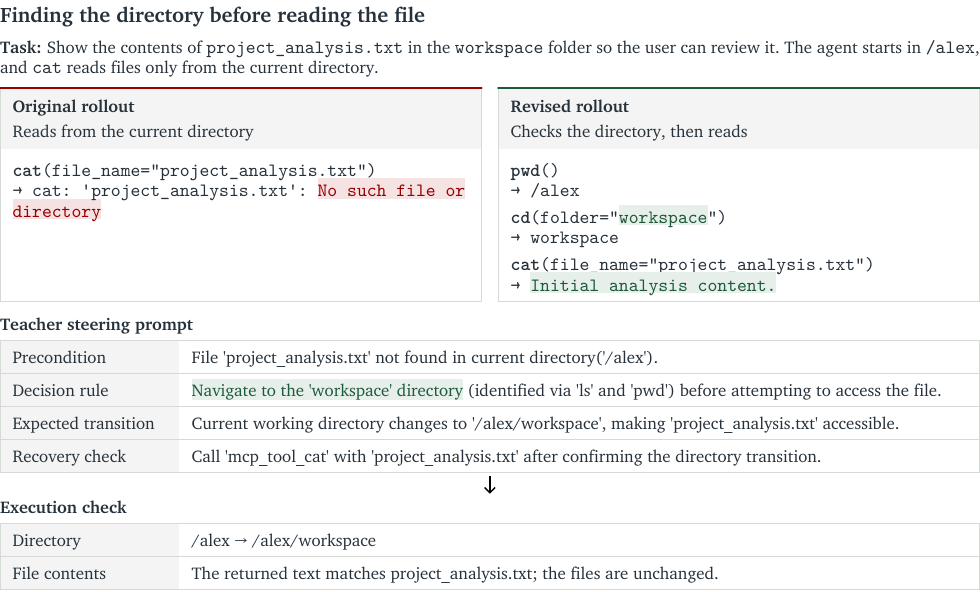}
\caption{Entering the workspace before reading the file. The reading tool
only accesses files in the current directory, so the unguided attempt to
read the workspace file from \texttt{/alex} fails. Under teacher steering,
the student checks its location, enters
\texttt{workspace}, and retrieves the requested file contents.}
\label{fig:app_training_directory}
\end{figure}

The training cases show how teacher steering helps the student correct a
failed decision from the pre-error history and state.
In the file-writing and directory-navigation cases, the guidance makes an
unmet prerequisite explicit: create the file before writing or enter the
workspace before reading
(\cref{fig:app_training_sentence,fig:app_training_directory}; see also
\cref{fig:overview_new}).
In the tool-list and card-control cases, the guidance clarifies which tools
and parameters to use. The student calls the advertised tool names without
their list numbers or sets the flag that enforces the country restriction
(\cref{fig:app_training_tool_list,fig:app_training_country}).

In the file-writing and tool-list cases, local verification checks the
result of the corrected action: the newly created empty file or the
unlocked doors match the reference state
(\cref{fig:app_training_sentence,fig:app_training_tool_list}).
The student then continues without the steering prompt, writing the
requested sentence or turning on the headlights.

\FloatBarrier
\subsection{Steering prompt comparison}
\label{app:steering_case}
\suppressfloats[t]

\begin{figure}[htp]
\centering
\includegraphics[width=\linewidth]{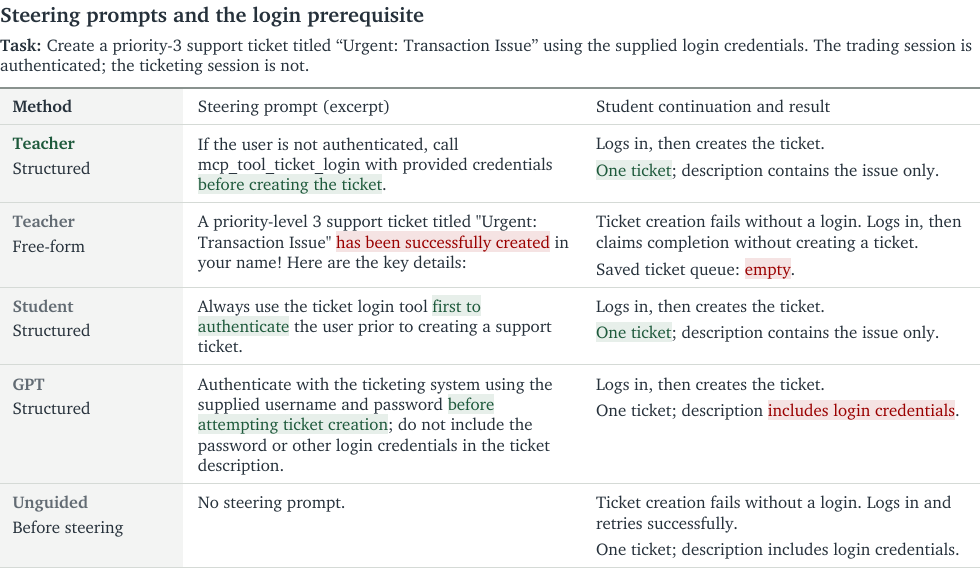}
\caption{Guidance content and student execution in ticket creation.
The four guided continuations start from the same history and environment
state. Structured teacher and student prompts lead to tickets without
credentials in their descriptions, while the free-form teacher prompt leads
to no ticket. Both the GPT-guided continuation and the original unguided
rollout create a ticket containing credentials; the latter first recovers
from a login error.}
\label{fig:app_ablation_combined}
\end{figure}

The ticket-creation case complements Q3 by showing how prompt format and
source affect the guidance the student receives
(\cref{tab:ablation_summary,fig:app_ablation_combined}).
Although the trading session is already authenticated, creating a ticket
requires a separate login. Structured prompts from both the teacher and the
student identify this prerequisite, and both continuations create the
ticket after login. The free-form teacher prompt instead claims that the
ticket already exists; the student logs in but stops without creating one.

The GPT-generated structured prompt also gives the correct login order and
instructs the student to keep credentials out of the ticket description.
The student logs in before creating the ticket but still includes the
credentials in its description. This partial adherence illustrates why
closed-loop teacher steering uses execution feedback to assess whether the
intended behavioral correction has occurred.

\FloatBarrier
\Needspace{275pt}
\subsection{Recovery cases}
\label{app:recovery_cases}
\suppressfloats[t]

\begin{figure}[htp]
\centering
\includegraphics[width=\linewidth]{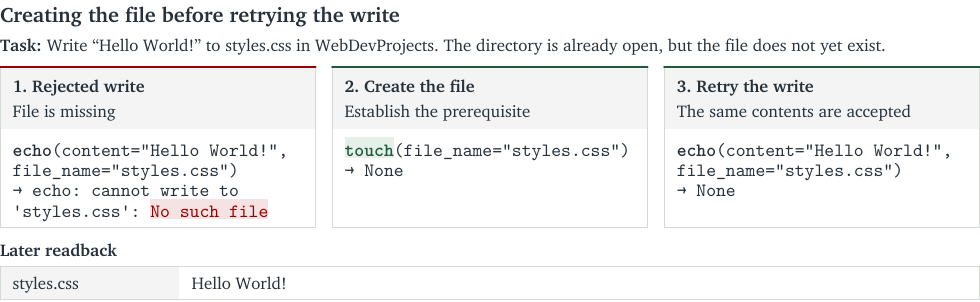}
\caption{Recovering from a missing-file error. \nickname creates
\texttt{styles.css} and retries the original write.
A later read returns the requested contents.}
\label{fig:app_recovery_create}
\end{figure}

\begin{figure}[htp]
\centering
\includegraphics[width=\linewidth]{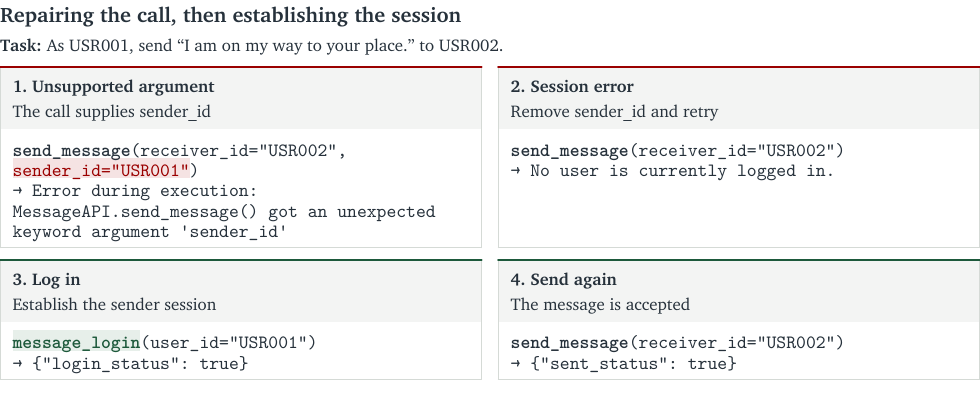}
\caption{Resolving successive errors before sending a message. \nickname
removes the unsupported \texttt{sender\_id} argument and retries. When the
revised call reports that no user is logged in, the agent logs in as the
sender and sends the message successfully.}
\label{fig:app_recovery_message}
\end{figure}

These cases illustrate an emergent behavior beyond the corrections
explicitly targeted during training. Our trajectory revision constructs
corrections from the restored pre-error history and state; at inference,
the distilled student also recovers from execution errors within its
ongoing rollout, without teacher guidance.
In the file-writing trajectory from \cref{fig:app_file_writes}, it creates
the missing file after a rejected write and retries
(\cref{fig:app_recovery_create}). In the messaging task, it successively
resolves an unsupported argument and a missing login session before sending
the message (\cref{fig:app_recovery_message}).

\FloatBarrier
\Needspace{9\baselineskip}
\section{Limitations}
\label{app:limitations}

Teacher-steered trajectory revision requires successful teacher references,
restoration of the pre-error history and environment state, and reliable
local verification. Without a successful reference, a failed rollout is
retained for distillation but receives no teacher-steered corrections.
Applicability to environments without state restoration or reliable local
verification remains untested. Moreover, local verification checks a revised
response's effect in the current context; it does not guarantee eventual
task success or establish the correctness of every reasoning step in the
response.

\end{document}